# Developing automatic verbatim transcripts for international multilingual meetings: an end-to-end solution

**Akshat Dewan**
**Michal Ziemski**
**Henri Meylan**
**Lorenzo Concina**
**Bruno Pouliquen**                    **{firstname.lastname}@**wipo.int
World Intellectual Property Organization, Global Databases Service
34, chemin des Colombettes, CH-1211 Geneva 20, Switzerland

**Abstract**

This paper presents an end-to-end solution for the creation of fully automated conference meeting transcripts and their machine translations into various languages. This tool has been developed at the World Intellectual Property Organization (WIPO) using in-house developed speech-to-text (S2T) and machine translation (MT) components. Beyond describing data collection and fine-tuning, resulting in a highly customized and robust system, this paper describes the architecture and evolution of the technical components as well as highlights the business impact and benefits from the user side. We also point out particular challenges in the evolution and adoption of the system and how the new approach created a new product and replaced existing established workflows in conference management documentation.

## 1. Introduction

This paper describes the experience of implementation of a system to automatically create post-factum meeting transcripts and their translations in various languages at WIPO. WIPO is a United Nations (UN) agency in charge of intellectual property and hosts many meetings over the calendar year[1]. These meetings are simultaneously interpreted into the six official UN languages – Arabic, Chinese, English, French, Russian and Spanish (AR, ZH, EN, FR, RU, and ES). Originally, for these meetings, English verbatim reports and their translations were all hand produced, resulting in high cost and delay. Now we deliver automatic transcripts and translations shortly after the end of the meeting.

All automatically produced transcripts and their translations are now published on a newly created public web portal[2], which combines live video, video-on-demand, S2T, MT, and a comprehensive search engine. It is integrated with other WIPO conference room systems, there-fore has access to automatic download of media files and meta data which allows for a rich user experience. The solution has also been already adopted by other international organizations.

---

[1] WIPO has more than 100 formal meeting days a year

[2] https://webcast.wipo.int/

The backend systems are customized using WIPO data and data gathered from the collaboration of various international organizations. Both neural S2T and neural machine translation (NMT) support all six UN languages and provide transcripts and translations in after-the-fact mode. The MT and S2T components are customized to work well for the language used in the meetings domain of international organizations. Our solution, based on open-source tools, is installed on-premises, allowing us to meet our strong data security and privacy policies, and is even fit for our confidential meetings. We took a conscious decision to cascade S2T with MT instead of end-to-end speech-to-translated-text approach, because it did not meet minimum quality requirements across all language pairs.

Overall, the new solution was well received by the users, creating a new type of product, and replacing existing manual workflows. The users accepted the trade off between potentially lower transcript accuracy for significantly reduced turnaround time. Collected user feedback regarding speed, access and quality shows a positive impact and a transformation of the way transcripts are used. Evaluation included automatic metrics like Word Error Rate (WER) and BLEU (Papineni, 2002) for S2T and MT and other more business-oriented metrics such as fitness for purpose, turnaround time, user experience and cost savings. Further work would be in improving transcript quality, increasing the scope of supported languages, and investigation of open questions such as preference and evaluations of different pathways to arrive at a certain transcript language: e.g., S2T on monolingual interpretation channel vs S2T + MT cascade on multilingual direct floor channel.

## 2. Background

WIPO, as a specialized UN agency, hosts many international, multilingual meetings each year. Speakers can deliver their interventions in any of the official languages of the UN or in Portuguese. The original speech is then simultaneously interpreted into the remaining official languages.

For all WIPO meetings, originally, verbatim reports were prepared in all UN official languages manually. These were very high quality reports but were costly and time-intensive to produce - anywhere from a month to sometimes 6 months.

With the help of our solution, which is a cascade of an S2T (WIPO S2T[3]), and an MT system (WIPO Translate[4]), we provide machine-generated transcripts and their corresponding machine translations in a couple of hours after the conclusion of a meeting. After running the system in pilot mode for one year for two of its especially important meetings, the WIPO General Assemblies for most of its meetings adopted this system, and it has replaced the manually prepared verbatim reports. This includes the processing of internal confidential WIPO meetings.

Other international organizations are also building speech to text solutions for various use cases, and we are closely collaborating with some. The European Parliament and European Commission are building systems to make their meetings more accessible to a wider audience. The United Nations office of Geneva, International Labour Organization, World

---

[3]https://www.wipo.int/about-ip/en/artificial_intelligence/speech_to_text.html

[4]https://www.wipo.int/wipo-translate/en/

Trade Organization, and the European Union Court of Justice are also leveraging the WIPO technology to accelerate their work of report writing.

## 3. Motivation and Design

WIPO's extensive experience in the development of innovative MT models was a major driver behind the efforts to explore S2T for WIPO's needs. When we started our exploration in 2018, we trained an English model from scratch and benchmarked it against other commercial and open-source (Amodei, 2015), (Collobert, 2016) providers on our in-domain test set. We obtained competitive overall accuracy, which along with other business needs for terminology accuracy, turnaround time and strong confidentiality requirements, motivated us to continue developing in-house and guided our design choices.

In 2018, hybrid HMM DNN Kaldi recipes were state-of-the-art (Hadian, 2018), but we chose to use the end-to-end approach using RETURNN (Zeyer, 2018). Our decision relied on the promise (Chan, 2016), (Chiu, 2017), (Doetsch, 2016) of the attention based RNN models and the relative ease of training models in the end-to-end paradigm.

We recognized the fast pace of evolution of the state of the art and kept a modular architecture to easily integrate other providers in our pipeline. Currently, we have RETURNN and ESPnet integrated in our production pipeline. ESPnet[5] Transformer models provide more accurate (lower WER) results compared to RETURNN RNN models. On the other hand, RNN models outperform Transformer models on the metric of adequacy measured in terms of deletion errors.

We can select the model with the highest fitness for purpose. We are also planning to add support for various commercial S2T providers to further increase our options.

## 4. Training data

A major hurdle for in-house S2T development was obtaining in-domain training data, as the originally collected internal WIPO data was limited in size, especially in languages other than English. To overcome this challenge, we did the following:
1) we collaborated with other international organizations to leverage their historical meetings data,
2) contracted external providers to prepare transcriptions of WIPO in-domain audio,
3) and bought out-of-domain proprietary corpora

**Collaboration with International Organizations**: As international organizations hold many meetings every year, we targeted meeting audios and their corresponding reports for our needs. Obtaining such data and using it for training poses several hurdles. Using interpretation audio has a potential intellectual property right based legal issue. There are other administrative and technical obstacles before such data becomes available and usable.

*Sourcing and filtering*: Very often, meeting audio and meeting reports are owned by different business units and thus are stored separately. This can lead to poor mapping between meeting audio and report files. In many cases, the exact type of content varies, and reports can be heavily edited and contain a lot of indirect speech. We filter out such reports to reduce the amount of noise in training data. For text extraction, we have to deal with meeting records that are kept in several different document formats such as Doc, Docx, RTF, and PDF. Extracting text from these documents needs special care or it can lead to noisy data.

---

[5] https://github.com/espnet/espnet

*Long-term storage and limitations of use*: We must also take special care when the partner data is sensitive and not publicly available. E.g., making special arrangements for isolating long-term storage for such data and limiting the use of models trained with such data.

**Crowdsourcing transcripts**: We also contracted external providers to create some in-domain training data transcripts. Such transcripts were cheaper but suffered from inconsistencies and errors in WIPO terminology. We had to run several iterations of harmonization and cleanup to reach the desired level of accuracy. We also leveraged the Arabic and Chinese speakers at WIPO to collect audio for some of our glossaries and terminology databases.

**Commercial corpora**: Apart from freely available public corpora – librispeech (Panayato, 2015) (Pratap, 2020), common voice[6], multitedx (Salesky, 2021), m-ailabs[7], open STT[8] etc, we also procured some commercially available corpora for Arabic and Chinese as can be seen in figure 1.

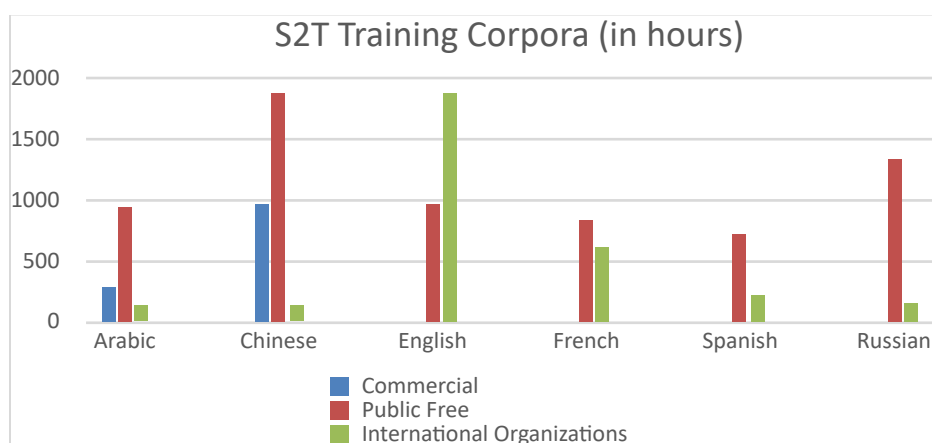

**Figure 1: Breakdown of the training corpora for six UN languages in hours of audio.**

---

[6]https://labs.mozilla.org/projects/common-voice

[7]https://www.caito.de/2019/01/03/the-m-ailabs-speech-dataset/

[8]https://github.com/snakers4/open_stt

## 5. Training S2T

In this section, we describe our pipeline (see Figure 2) to obtain training corpora from the sources data mentioned in section 4.

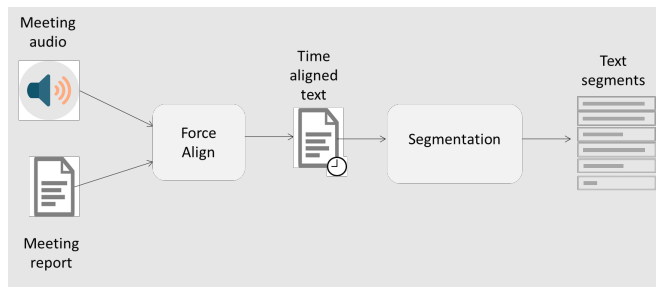

**Figure 2: The training pipeline**

**Audio text alignment /Word level alignment**: As each meeting audio file runs for hours and is not compatible with present-day neural architectures, it must be split into segments. To that end, text extracted from meeting reports must be aligned with meeting audio. Gentle[9], with a publicly available Kaldi model is used for English, however, as such models are not readily available for other UN official languages, we trained Kaldi models for ES, FR, ZH, AR and RU. This results in word level alignment of text and audio.

**Segmentation / Sentence level alignment**: We create training segments using word level timing information obtained from forced alignment. We carry out three distinct kinds of segmentations: 1) linguistic segmentation – based on sentence boundaries; 2) length-based segmentation; and 3) a hybrid of the 1) and 2).

**Data augmentation:** We use speed perturbation of audio to augment our training data. We also add silence and music examples in the training data with the goal of increasing robustness.

**Pre-processing text:** We tokenize the text and replace cased text with casing tags to decrease the vocabulary size. In addition, the multilingual nature of our meetings sometimes leads to crosstalk between different channels; therefore, we annotate the foreign language audio with special tags. The training corpora thus prepared are combined, harmonized, and filtered before proceeding for training different RNN, and Transformer models using RETURNN and ESPnet training scripts.

## 6. Machine translation

We chose to cascade S2T with MT to provide speech to translated text. One reason for this choice was insufficient training data for end-to-end speech translation and another reason was to leverage highly performant MT models for the meetings domain of WIPO. These WIPO Translate MT (Pouliquen, 2017) models are custom-trained using text from WIPO meetings and other documents.

---

[9] https://github.com/lowerquality/gentle

To further customize the MT models for cascading with S2T, we experimented with two methods: 1) denoising models that improve the quality of automatic transcriptions and make them more like the input expected by an NMT system; 2) noising models that replicate errors produced by the S2T system and can be used to introduce errors to the source side of the parallel MT corpora to make the MT systems more robust to automatic transcriptions.

Our initial noising and denoising experiments did not allow us to efficiently model the transcription errors; therefore, we did not further pursue this approach.

## 7. Decoding pipeline

In this section, we describe our decoding pipeline. Figure 3 illustrates various inputs and outputs of our system.

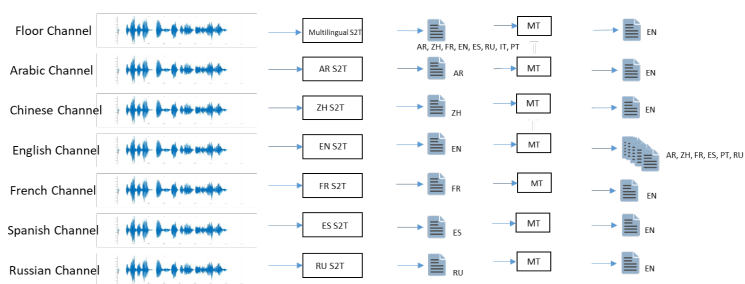

**Figure 3: Inputs and outputs**

**Input videos**: As seen in Figure 3, WIPO meetings have 7 audio channels – 6 for the 6 official UN languages plus 1 for the floor. The floor channel is multilingual and has the speaker's original audio while the other 6 channels are monolingual and may contain interpretation. When the language of a monolingual channel matches the language of the multilingual floor channel, the monolingual channel contains the same audio as the multilingual floor channel; otherwise, the monolingual channel has the audio from the interpretation booth.

**Meta data**: A tight integration with other WIPO systems allows us access to rich meta data to enhance the user experience. The metadata consists of the following items:
- Meeting title and category
- Agenda items and their corresponding timestamps
- Speakers and their corresponding timestamps (this information is more accurate and informative than automatic speaker diarization)
- Speaker information (Flag, Biography etc.)
- List of associated meeting documents

We have a fully automatic pipeline that fetches the meeting video files and metadata from the media server as soon as a meeting concludes.

**Output**: Outputs constitute S2T transcripts for the 7 audio channels followed by machine translations of the S2T transcripts. Once the media and meta data are available, we run S2T, and MT systems followed by indexing everything in Elasticsearch[10]. Finally, the video, tran-scripts and the translations are published on our public facing webcasting portal. The following bullet points summarize the workflow followed for each of the 7 audio channels:
- Split the several hour-long audio file into 1-20 second segments. We use webRTC[11] or Silero[12] for VAD, and Kaldi[13] for segmentation.
- The segments are then fed to WIPO S2T to generate text, which is then filtered and normalized.
- The English S2T generated text is machine translated into the five remaining UN official language and Portuguese. S2T generated texts in other languages are machine translated back into English (including the multilingual floor channel where each sentence is either copied, if originally in English, or translated into English).
- The audio and the normalized text are force aligned to obtain word level alignment between text and audio.
- Meeting metadata along with the machine transcripts are then indexed in Elasticsearch, which enables a powerful search functionality.
- Some users prefer the rich functionality and familiarity of the Word format to our web interface, so we create exports in a Docx format at the last step.

---

[10] https://www.elastic.co/

The output files are uploaded to the webcasting portal[14] as they become available. Typically, this means that the English transcript is online before the other languages.

## 8. Webcasting Portal

Before 2022, WIPO had two separate websites – one for WIPO S2T and another for meeting live video and VODs. This was not ideal and with the goal of enhancing experience for WIPO stakeholders, we designed and developed a new unified web portal: webcast.wipo.int

The portal offers S2T text that helps in making WIPO meetings content become more available and visible. In addition, it becomes accessible to the hearing impaired. While WIPO meetings are multilingual to start with, the portal further cuts the linguistic barrier by providing machine translation of the S2T generated text. This allows e.g. a non-Chinese user to read the English translation of the original statement made by the Chinese delegation in Chinese. We have also taken particular care to make the portal accessible to the visually impaired by having screen reader friendly web design. The WIPO meetings meta data along with WIPO S2T text allows for smooth navigation and search across the content. The powerful search enables content discovery across the full WIPO meetings library, including records from the past.

**Systems Integration:** The webcasting portal was envisaged to have a tight integration with other IT systems at WIPO. E.g., the webcasting portal interacts with the videoconferencing

---

[11] https://github.com/wiseman/py-webrtcvad

[12] https://github.com/snakers4/silero-vad

[13] https://kaldi-asr.org/

[14] https://webcast.wipo.int

software system from Arbor Media called Connectedviews[15] via a web API. Connectedviews itself is integrated with other videoconferencing systems like Televic[16] that provide meta data information like "who spoke when." The WIPO Diplomatic Engagements Team also adds other metadata to the Connectedviews system. This integration reduces the turnaround time of meeting transcripts for our stakeholders. We encountered several hurdles during integration, and some are described here:

*Inconsistency in meeting titles*: Meeting titles are manually entered in the videoconferencing software, and this can cause some inconsistencies. These inconsistencies could lead to duplicate entries on our public facing portal. To avoid that, we set up a strict naming convention for meeting titles. This, however, does not mitigate all the risks of having duplicates in the web portal, therefore, we need to rely on effective communication between teams to overcome this challenge.

*Changes to metadata*: Since some meta data is dynamic[17], we have opted for a polling method to update the meta data on our portal continually.

*Communication*: Continuous communication with the different stakeholders is an important aspect of our work. It is especially important because of the multitude of ways in which the portal can be used. One example is as follows: we decode the 6 monolingual channels using monolingual S2T models while we decode the multilingual floor channel using a multilingual S2T model. Monolingual audio channels contain the same audio as the multilingual floor channel when floor language is the same as the monolingual channel language. This leads to a scenario where one audio has two different transcriptions. This causes confusion for the users, and we solve this by one-on-one user training sessions.

## 9. Experiments with Whisper

We have also been experimenting with the OpenAI Whisper (Radford, 2022) models. For now, we have focussed on using Whisper only for S2T and we plan to investigate the translation feature in the future. We are investigating customization of pre-trained models for improving performance on in-domain terminology. Customization paths that we currently investigate are:

- Fine-tuning pre-trained models using in-domain data from WIPO and other international organizations.

- Prompting strategies for the pre-trained models to improve performance for WIPO terminology

In our fine-tuning experiments, especially for high-resource languages like English and French, fine-tuning improved the recognition performance on in-domain terminology, but the general accuracy remained unchanged or even slightly worsened. These experiments suggest that it would be difficult to improve performance for high resource languages. However, we plan to run more fine-tuning experiments for languages like Arabic, Chinese and Russian. Whisper models have unpredictable behaviour for multilingual audios, and we are trying to use speaker diarization to resolve that. We are also investigating prompting strategies (Radford, 2022) to improve in-domain recognition accuracy. Early prompting experiments in English have shown promising but sometimes unpredictable results and we are still working on perfecting our strategies.

---

[15]https://www.arbormedia.nl/products/connected-views

[16]https://www.televic.com/en/conference/markets/institutions

[17] For example, country names can change – e.g., Turkey was recently changed to Türkiye

## 10. Evaluation and benchmarking

We evaluate our S2T models on the widely used automatic evaluation metric of WER (and Character Error Rate - CER). Since it penalizes all errors equally, we also evaluate in-domain terminology on a smaller, terminology heavy test set to get a better understanding of perceived WER.

| Language | WIPO | GCP | AWS | Whisper Small | Whisper Medium | Whisper Large |
|---|---|---|---|---|---|---|
| EN | 0.148 | 0.123 | 0.118 | 0.109 | 0.102 | 0.107 |
| FR | 0.056 | 0.171 | 0.102 | 0.149 | 0.094 | 0.085 |
| ES | 0.101 | 0.126 | 0.108 | 0.108 | 0.079 | 0.073 |
| ZH | 0.071 | 0.070 | 0.061 | 0.193 | 0.105 | 0.125 |
| RU | 0.145 | 0.255 | 0.319 | 0.278 | 0.253 | 0.238 |
| AR | 0.191 | 0.473 | 0.264 | 0.487 | 0.340 | 0.508 |

Table 1: WER (CER for ZH) values for various models and services in Dec 2022

We note that our test sets have varied sizes, e.g., EN has 3051 examples while FR has only 252 and this can introduce some bias in our evaluations. We also note that while both AWS and GCP have features to customize the models, we used their off-the-shelf models in our benchmarking.

We also performed human evaluation with the help of linguists to evaluate fitness for purpose where errors were classified as minor or disruptive. Disruptive errors delete or substitute important parts such as nouns, verbs, proper names, time, places, numbers etc. and affect text comprehension. Minor errors do not change the sentence's meaning, for example, errors in function words - pronouns, prepositions, adjectives, and adverbs.

We note that human evaluation for Whisper models and commercial providers has not yet been carried out and is a work in progress.

## 11. User Feedback

The WIPO General Assemblies (GA) reached a consensus in September 2019 to approve a S2T pilot. It was proposed as a replacement for the verbatim reports for two of nine WIPO bodies. We collected regular feedback during the pilot phase and after its successful completion, WIPO GA in 2021 approved the WIPO S2T for all but two of its meetings. While the users highlight the inaccuracies in certain languages other than English, the overall feedback is extremely positive. The rapid availability and the cost savings are congratulated. It also aligns with the Organization policy for increased digitization. Users also appreciate the user interface, which enables new and improved working methodologies.

## 12. Conclusions and future work

The deployment of the WIPO S2T and its consolidation into the publicly available WIPO webcast portal created a new type of meeting transcripts, which was well received both internally and externally. Even though the produced texts contain errors, according to user feedback, this was by far outweighed by the speed and convenience of presentation and availability – including various languages and accessibility. This switch also achieved a considerable cost reduction at WIPO.

Automatic workflows and a tight integration with existing systems has proven crucial to the success and adoption of the system. Working closely with the business side in short iterations, and particular focus on communicating the advantages and limitations of the solution have been critical to be able to achieve a positive reception of the final new product, shown by feedback collected.

While most models and components are trained in-house, it remains to be seen which mix of such components will deliver the best possible quality and experience for users in the long term. The modular architecture of the system allows reacting flexibly to user demands and ongoing developments. Especially for high resource languages such as English and French, pre-trained components may be more than competitive to the existing in-house produced. In the short term, we intend to improve in-domain terminology accuracy by using customized Whisper models.

For future work, we would like to make our benchmarking more comprehensive, by including more commercial S2T providers and using the customization features they offer. There are also many open questions: such as the best way to allow for and incorporate the manual edition of pieces of transcripts; and how to best use the multitude of various outputs generated from different audio streams: for example, direct S2T on interpretation channel vs MT cascade.

## Acknowledgements

The authors wish to thank Sandrine Ammann, Alessio Corsini, Thomas Gerdes, Roman Grundkiewicz, Sofia Lobanova, Christophe Mazenc, Husaini Mohammad, Andrew Moore, Ha Nguyen, Proyag Pal, Daniel Torregrossa, Tania Romera, Antonella Russo, Jeremy Thille, and Marie-Pierre Vincent for their contributions and invaluable assistance with the project.